\documentclass{article}
\usepackage{ijcai26}

\usepackage{times}
\usepackage{soul}
\usepackage{url}
\usepackage[hidelinks]{hyperref}
\usepackage[utf8]{inputenc}
\usepackage[small]{caption}
\usepackage{graphicx}
\usepackage{amsmath}
\usepackage{amsthm}
\usepackage{booktabs}
\usepackage{algorithm}
\usepackage[switch]{lineno}

\usepackage{enumitem}
\usepackage{amssymb}
\usepackage{algpseudocode}
\usepackage{multirow}
\usepackage{longtable}
\usepackage[normalem]{ulem}
\usepackage[page]{appendix}
\usepackage{subfig}

\title{GraphAllocBench: A Flexible Benchmark for Preference-Conditioned Multi-Objective Policy Learning}

\author{
Zhiheng Jiang$^1$
\and
Yunzhe Wang$^2$\and
Ryan Marr$^2$\and
Ellen Novoseller$^3$\and
Benjamin T. Files$^3$\And
Volkan Ustun$^2$\\
\affiliations
$^1$University of California, Los Angeles\\
$^2$USC Institute for Creative Technologies\\
$^3$U.S. Army DEVCOM Army Research Laboratory\\
\emails
zhjiang@g.ucla.edu,
\{wyunzhe, ustun\}@ict.usc.edu,
\{ellen.r.novoseller, benjamin.t.files\}.civ@army.mil
}

\begin{document}

\maketitle

\begin{abstract}
Preference-Conditioned Policy Learning (PCPL) in Multi-Objective
Reinforcement Learning (MORL) approximates diverse Pareto-optimal
solutions by conditioning a single policy on user-specified preferences,
enabling run-time adaptation to arbitrary trade-offs without retraining.
However, existing PCPL benchmarks are largely restricted to toy tasks and
fixed environments, limiting their realism and scalability. To address this
gap, we introduce GraphAllocBench, a flexible benchmark built on
CityPlannerEnv, a novel graph-based resource allocation sandbox inspired by
city management. GraphAllocBench provides a rich suite of problems with
customizable objective functions, varying preference conditions, complex
Pareto Fronts, and high-dimensional scalability. We further propose two
supplementary metrics -- Proportion of Non-Dominated Solutions (PNDS) and
Ordering Score (OS) -- that capture prediction reliability and preference consistency while
complementing the widely used hypervolume metric. Through experiments with
several state-of-the-art PCPL algorithms and our own MLP and graph-aware
PCPL-PPO baseline, we show that GraphAllocBench exposes distinct failure
modes that hypervolume alone does not capture but our supplementary metrics
reveal, while motivating graph-based approaches such as Graph Neural
Networks (GNNs) for scaling to complex, high-dimensional allocation tasks.
By letting users freely vary objectives, preferences, and allocation rules,
GraphAllocBench serves as a versatile and extensible testbed for advancing
PCPL.
Code: https://github.com/jzh001/GraphAllocBench
\end{abstract}

\begin{figure*}
    \centering
    \includegraphics[width=\textwidth]{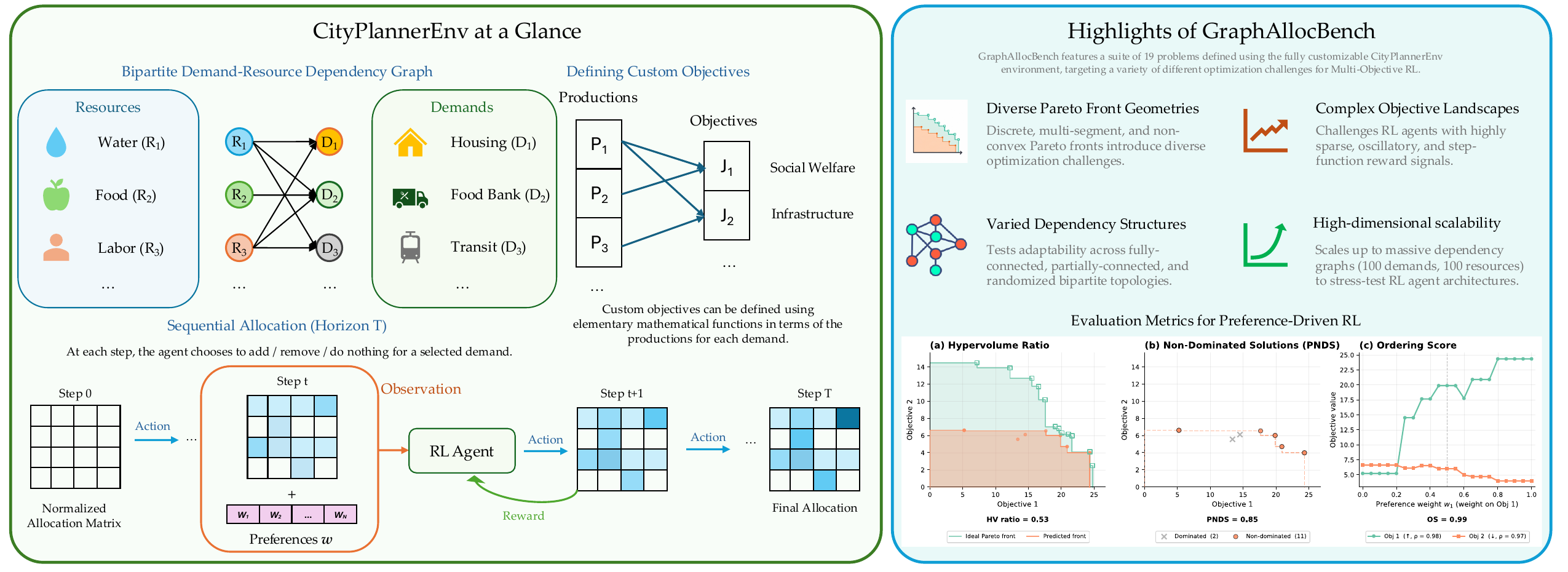}
    \caption{CityPlannerEnv models city resource allocation as sequential decision-making over a bipartite demand-resource graph, with composable multi-objective rewards and preference-conditioned policies. GraphAllocBench provides a diverse suite of problems spanning varied graph structures, Pareto front geometries, and scales, built on top of CityPlannerEnv.}
    \label{fig:summary_figure}
\end{figure*}

\section{Introduction}
Reinforcement learning (RL) has achieved remarkable progress in domains such as video games ~\cite{DBLP:journals/corr/abs-1912-10944}, robotics ~\cite{labiosa2025reinforcementlearningclassicalrobotics}, and large language model fine-tuning~\cite{ouyang2022traininglanguagemodelsfollow}. 
Much of this success is built on single-objective RL algorithms such as Proximal Policy Optimization (PPO) ~\cite{Schulman2017ProximalPO}, which optimize a single scalar reward function. 

However, many real-world and simulated environments involve multiple, often conflicting objectives rather than a single scalar reward. For example, in games and training simulations, Non-Player Characters (NPCs) must balance dealing damage, avoiding damage, and maximizing survival time ~\cite{schrum:aiide08} to create engaging and adaptive experiences. In such cases, objectives carry different levels of importance; we term such weights \textit {preferences} over objectives.

A common approach to multi-objective RL (MORL) is scalarization  ~\cite{6615007} (See Appendix \ref{subsec:scalarization})
, where objectives are collapsed into a single reward using preference weights. While straightforward, using a predetermined scalarization ties the learned policy to one fixed preference and requires retraining for new ones.

Other existing approaches to MORL (See Section \ref{subsec:morl}) include training populations of policies via evolutionary algorithms ~\cite{PBIDebAndJain,dixit2025preference}, particle swarm approaches ~\cite{becerril2025particle} or training multiple single-objective models to approximate the Pareto front ~\cite{xu2020prediction}. These methods are generally inefficient, as each new preference requires retraining from scratch, or requires keeping track of multiple concurrent policies. Gradient-based optimization with vector-valued rewards ~\cite{DESIDERI2012313} has also been explored, but often suffers from convergence issues ~\cite{NEURIPS2021_7bb16972}.

Preference-conditioned Policy Learning (PCPL) in MORL, for instance PD-MORL ~\cite{basaklar2023pdmorl}, offers a more scalable solution. Instead of training a separate model for each preference, a single preference-conditioned policy can adapt to arbitrary user-specified preferences. 
Compared to alternatives such as role embeddings  ~\cite{long2024roleplaylearningadaptive} or policy pools ~\cite{garnelo2021pickbattlesinteractiongraphs}, 
preference-conditioned methods provide a more direct and flexible mechanism for handling diverse objectives.

Despite algorithmic progress in the field, current multi-objective benchmarks~\cite{felten_toolkit_2023} are insufficient for pushing progress forward for PCPL. Most existing testbeds utilize either continuous optimization problems  \mbox{~\cite{Deb2005,WFG}} or simple 2D grid-based games ~\cite{envelope,deepseatreasure2011}. These lack flexibility in scaling observations, actions, objectives and dependency structures, making them insufficient to evaluate the robustness of preference-conditioned methods in more complex problems. For instance, in Deep Sea Treasure~\cite{deepseatreasure2011}, the objectives are to maximize treasure and minimize time, as limited by the game mechanics.

To address this gap, we introduce \textbf {CityPlannerEnv}, a Gymnasium-based ~\cite{towers2024gymnasium} sandbox environment inspired by city-scale resource allocation that enables multi-objective planning in graph-based environments of arbitrary and scalable complexity. Cities must allocate limited resources across diverse and often conflicting demands -- such as reducing congestion, fostering economic growth, and promoting sustainability -- according to shifting preferences. The environment models this process as a sequential allocation problem, represented through bipartite resource-demand dependency graphs. Agents must not only balance competing objectives but also anticipate future events and adapt to varying risk profiles (e.g., conserving resources for emergencies versus aggressively pursuing short-term gains). Although framed around city planning, the bipartite demand–resource structure is the abstraction underlying many allocation problems such as transportation, human resources and supply-chain distribution.

In \textbf{CityPlannerEnv}, 
agents allocate resources by adding or removing productions for each demand at each step to pursue custom-defined objectives based on user preferences. This incremental allocation mechanism draws inspiration from the Multi-step Colonel Blotto Game (MCBG), a well-studied model for competitive resource allocation on graphs ~\cite{an2025reinforcementlearninggametheoreticresource,an2025doubleoraclealgorithmgametheoretic,shishika2021dynamicdefenderattackerblottogame}. Dependencies between resources and demands are represented as bipartite graphs, and objectives can be defined over any subset of productions. Agents seek to plan actions that maximize the expected cumulative reward over time while respecting preference trade-offs. As the number of resources, demands, and objectives increases -- and as conditions change over time -- the allocation problem becomes increasingly 
challenging for PCPL.
 
We further introduce \textbf{GraphAllocBench}, a benchmark for PCPL that defines a suite of challenging problems designed using 
CityPlannerEnv. 
GraphAllocBench leverages the flexibility of CityPlannerEnv to create customizable problems by varying the numbers of demands, resources, objectives, and objective shapes, providing targeted stress tests for PCPL algorithms. 
For example, GraphAllocBench includes test problems spanning diverse optimization challenges, including difficult objective functions, diverse and non-convex Pareto fronts, sparse observation spaces, complex dependency structures, and high-dimensional graph observations. For performance characterization, the benchmark includes the hypervolume ratio along with two novel supplementary metrics -- Proportion of Non-Dominated Solutions (PNDS) and Ordering Score (OS) -- to assess how well policies satisfy varied preferences. Figure \ref{fig:summary_figure} shows an overview of CityPlannerEnv and GraphAllocBench. 

To establish baseline performance and highlight how the benchmark challenges existing methods, we demonstrate a PPO-based ~\cite{Schulman2017ProximalPO} PCPL solution (PCPL-PPO) with Multi-Layer Perceptron (MLP) and Heterogeneous Graph Neural Network (HGNN) ~\cite{DBLP:journals/corr/abs-2003-01332} feature extractors. We further employ Smooth Tchebycheff Scalarization ~\cite{lin2024smoothtchebycheffscalarizationmultiobjective} to approximate challenging non-convex Pareto fronts and demonstrate how graph neural networks can learn preference-conditioned, graph-aware policies in environments with complex dependency structures, going beyond prior work that primarily focused on 2D grids ~\cite{basaklar2023pdmorl} or simpler graph topologies ~\cite{lin2022pareto,10766498}.\\

\noindent\textbf{Our contributions are as follows:}
\begin{itemize}[topsep=3pt]
    \item \textbf{GraphAllocBench benchmark:} We introduce \emph{GraphAllocBench}, a flexible and challenging graph-based benchmark for preference-conditioned policy learning, based on our novel, customizable sandbox environment \emph{CityPlannerEnv}, designed to capture complex resource allocation scenarios with complex bipartite dependency structures, discrete non-convex Pareto Fronts, and difficult objective functions.
    
    \item \textbf{Novel evaluation metrics:} Beyond the standard hypervolume metric, we introduce the \emph{ordering score} (OS) and the \emph{proportion of non-dominated solutions} (PNDS) to evaluate preference-consistency and robustness, addressing limitations of existing multi-objective evaluation practices for PCPL algorithms (see Section \ref{sec:related-work}).
    
    \item \textbf{PCPL-PPO with MLPs and HGNNs:} We use MLP-based PCPL-PPO as a strong baseline and introduce a preference-conditioned HGNN-based PCPL-PPO that substantially improves hypervolume coverage on high-dimensional graph observations, while exposing a measurable trade-off between coverage and per-objective preference alignment that remains an open challenge.

\end{itemize}

\section{Background} \label{sec:related-work}

\subsection{Multi-Objective RL} \label{subsec:morl}

Many solutions in the multi-objective optimization space have aimed to tackle the issue of learning Pareto Fronts (See Appendix \ref{sec:pareto-front}).
Evolutionary methods such as NSGA-III ~\cite{PBIDebAndJain} have shown success, focusing on generating increasingly Pareto-optimal populations of solutions with each generation by performing operations such as selection, crossover and mutation. However, while these evolutionary methods are effective for simpler multi-objective optimization tasks, they are less adaptable towards more complex and high-dimensional tasks, such as graphs, and struggle to maintain a set of diverse solutions ~\cite{GU2022117738}. 

More importantly, popular multi-objective optimization algorithms including evolutionary methods often aim to produce a set of solutions on the Pareto Front, making it difficult to specify a preference for a particular solution directly as input. Learning and maintaining a large population of Pareto-optimal solutions can be computationally and memory intensive, especially for higher-dimensional objective spaces. Hence for a large number of objectives, it becomes difficult to maintain a good resolution of different solutions on the Pareto Front. 

For Pareto-Optimal Policy Set Learning in RL, Multiple-Gradient Descent Algorithms (MGDA) ~\cite{DESIDERI2012313} have been introduced to approximate the Pareto Front. These algorithms involve iteratively updating variables such that all objectives are improved at the same time. However, these methods do not allow us to easily specify preferences for solutions with different tradeoffs,  and face convergence issues for some segments on the Pareto Front ~\cite{NEURIPS2021_7bb16972}.

Some work has been conducted to learn preference-aware Pareto-optimal policies using RL. One such approach is to train multiple single-objective RL models on a set of fixed preference vectors, and select a corresponding model to use during inference, based on a specified preference. A variant of this method uses an evolutionary algorithm with a population of different RL policies ~\cite{xu2020prediction}. A limitation of this method is that we can only learn a discrete set of preferences, as we need to train a separate RL model for each fixed preference vector. Additionally, these methods struggle with similar challenges to traditional multi-objective optimization tasks, as they still involve generating a population of Pareto-optimal solutions instead of an individual preference-aware solution.

Unlike conventional multi-objective approaches that generate a complete Pareto Front, Preference-Conditioned Policy Learning (PCPL) methods ~\cite{basaklar2023pdmorl,lin2022pareto,liu2025pareto,envelope} use a continuous, user-defined preference as input to a single policy. This allows them to produce a specific, preference-aware solution on the Pareto Front via a series of sequential decisions within the environment. The final reward can be computed using the scalarization (See Appendix \ref{subsec:scalarization})
of the components of the objective function, based on the preference provided during input. This allows us to learn a preference-conditioned policy, where during inference, we can specify any preference vector within a continuous range and obtain an approximation of the Pareto-optimal solution on the Pareto Front. 

\subsection{Benchmarking MORL}

Several benchmarks have been proposed for multi-objective optimization and reinforcement learning. For multi-objective optimization problems, examples include DTLZ ~\cite{Deb2005} and WFG ~\cite{WFG}, which are primarily traditional multi-objective optimization problems, where we are varying decision variables to approximate mostly continuous objective functions and Pareto Fronts. These benchmarks offer expandable and flexible testcases for multi-objective optimization. However, these benchmarks are not easily adaptable for MORL tasks involving sequential decision making, non-continuous Pareto Fronts, or high dimensional observation spaces such as graphs.

Common benchmarks for MORL include game environments, for instance those in Multi-Objective Gymnasium ~\cite{felten_toolkit_2023} such as Deep Sea Treasure (DST) ~\cite{deepseatreasure2011} and Fruit Tree Navigation (FTN) ~\cite{envelope}. These are game environments, e.g. where an agent tries to accomplish multiple objectives in a 2D grid world or MuJoCo task with continuous action spaces ~\cite{zhu2023paretoefficient}. 
However, these benchmarks are still limited to the mechanics of the game environment itself (limited action and observation spaces), and it is difficult to design custom scenarios and objective functions to challenge the RL model. Even though variants of these problems ~\cite{ReviewOfDST} have also been introduced to increase the size of the observation space and number of objectives, these variants are still non-expandable and normally focus on small action and observation spaces, with limited number of objectives, often just 2-3 objectives ~\cite{zhu2023paretoefficient}. 

GraphAllocBench addresses these gaps along three axes that existing benchmarks lack: (1) customizable objectives, enabling deliberate design of Pareto Front geometry; (2) scalable bipartite graph structure, exposing the limits of MLP-based approaches; and (3) complementary evaluation metrics (PNDS and OS) that go beyond hypervolume: PNDS measures prediction reliability, while OS is a preference-specific metric that measures preference alignment.


\section[CityPlannerEnv]{\NoCaseChange{CityPlannerEnv}}
Our sandbox environment \textbf{CityPlannerEnv} is modeled as a city that has a set of resources $R$ and a set of demands $D$ (Figure \ref{fig:summary_figure}). At each step, the RL agent receives the current resource allocation state as an observation, takes an action to alter the allocation, and receives a multi-objective reward signal.

\subsection{Definitions}

\begin{description}
    \item[Resource] This refers to a particular factor of production, 
    denoted by $R_i$, which can be combined with other resources to create productions (e.g. water, food, labor). The environment starts with a predefined number of units for each resource.
    \item[Demand] A type of need fulfilled by resources (e.g. food bank), denoted by $D_i$.
    \item[Requirements] Unweighted dependency set $R'(D_i) \subseteq R$ of the resources needed by each demand $D_i$.
    \item[Allocation] Assignment of resources to demands, represented by allocation matrix $A$, where $A_{i, j}$ is the amount of $R_i$ given to $D_j$. The observation returned by the environment is the normalized allocation matrix, concatenated with the current preferences vector for each rollout.
    \item[Production] Actual output for each demand, determined by the limiting allocated resources. Production $\mathbf{P}$ changes when an action is taken to add or remove a production, where $|\mathbf{P}|$ is the number of demands.
    \item[Objective Function] Multi-objective reward signal at the end of each step (i.e. after each action), defined by $\mathbf{J}(\mathbf{P}): \mathbb{R}^{|\mathbf{P}|} \to \mathbb{R}^N$, mapping $|\mathbf{P}| = |D|$ productions to $N$ objective values.
    \item[Preferences] Importance assigned to each objective, represented by preferences vector $\mathbf{w} = [w_1, \dots, w_{N}], \; \sum_{n=1}^{N} w_n = 1$, $0 \le w_i \le 1$ for each $i \in \{1, \ldots, N\}$.
\end{description}

\subsection{Action Space}
Custom action spaces can be defined within the CityPlannerEnv environment to move resources between demands. In our solution, we define a single action as the selection of 2 sub-actions. Firstly, the agent will choose to either add or remove a production or not to modify the current allocation state. Secondly, the agent will choose a specific demand among the demands $D$ for which to create 1 unit of production, remove 1 unit of production, or do nothing, depending on the first action chosen. When we add or remove a production, the corresponding resources tied to that demand will also be removed or added respectively from the unallocated resources pile.

CityPlannerEnv is implemented using Gymnasium ~\cite{towers2024gymnasium}. Custom environment configurations can be designed for the number of resource and demand types, the number of available resources for each type of resource, resource-demand dependency graphs, and objective functions using common mathematical functions (e.g. polynomial, sinusoidal, logarithmic). 

\subsection{Custom Objectives}
Objective functions in CityPlannerEnv are fully user-definable: each component $J_i$ can depend on any subset of productions $P_j$ and can be composed from elementary functions (linear, quadratic, logarithmic, logistic, Gaussian, ceiling, and floor, or sums thereof). This composability allows practitioners to deliberately engineer Pareto Front geometries, including smooth convex fronts, discontinuous multi-segment fronts, and non-convex fronts with oscillatory structure. Table \ref{tab:obj-fun} in Appendix \ref{sec:obj-fun} illustrates representative objective functions spanning these cases. 
This design flexibility is a key differentiator from existing benchmarks, where objectives are fixed by game mechanics.

\section[GraphAllocBench]{\NoCaseChange{GraphAllocBench}} \label{subsec:graphallocbench}

\textbf{GraphAllocBench} comprises a set of 19 challenging evaluation problems designed using our sandbox environment CityPlannerEnv, across six challenge
categories: Problem~0 serves as a baseline with a smooth,
convex Pareto front; Problems~1a--1c introduce oscillatory
and multi-segment objective landscapes with sparse rewards;
Problems~2a--2c increase the number of demands, expanding
the action and observation spaces; Problems~3a--3b scale to
five objectives with highly sparse reward signals;
Problems~4a--4b use non-fully-connected dependency graphs;
and Problems~5a--5e feature randomly sampled dependency
structures and objective functions at extended planning horizons.
Three large-scale problems (6a--6c) further stress-test scalability
on graphs with 100 demands.
Full problem specifications are provided in Appendix \ref{sec:obj-fun} (Table \ref{tab:problem-descriptions}).

This benchmark evaluates the quality of solutions using the hypervolume (HV) and Proportion of Non-Dominated Solutions (PNDS) using preferences sampled with the Das and Dennis method ~\cite{DasAndDennis1998}. The solutions evaluated via the Ordering Score (OS) are sampled separately (See Algorithm \ref{alg:ordering-score} in Appendix \ref{sec:ordering-score}). 

For each sampled preference, we roll out the agent's policy deterministically for $T$ steps to obtain the final allocation of resources to demands. The Das and Dennis method ~\cite{DasAndDennis1998} generates a set of evenly distributed preferences by constructing a structured grid of points on an $N$-dimensional simplex (for $N$ objectives). Hence, the number of preferences sampled increases combinatorially with $N$. For scenarios where it is computationally feasible to compute the allocations that lead to solutions on the analytical Pareto Front, we compare the hypervolume of the set of predicted solutions for the sampled preferences with the ideal Pareto Front. The evaluation metrics are defined respectively in the following sections.

\subsection{Hypervolume} \label{subsubsec:hypervolume}

For each sampled preference $\mathbf{w}$, we obtain a final allocation of resources, which allows us to calculate the final production vector $\mathbf{P}$ for a given allocation.  Hence, we can use the final production state to calculate the value of each objective component, which are collectively represented by $\mathbf{J}(\mathbf{P})$. By varying our preferences, we obtain sets of objective vectors $\mathbf{J}$, which form a set of predicted optimal solutions (i.e. the predicted Pareto Front). We can evaluate this solution set using the hypervolume metric based on the maximization variant of the PyMOO ~\cite{pymoo} implementation (See Appendix \ref{sec:hv-appendix}).

For simpler solutions, we can also calculate the ideal hypervolume of the environmental setup by enumerating all possible allocations. We can then calculate how well the predicted Pareto Front estimates the true Pareto Front using $\text{HV}_{\text{ratio}} = \text{HV}_\text{predicted}/\text{HV}_{\text{ideal}}$.

$\text{HV}_\text{ratio}$ measures agent performance against the best possible solution and, being normalized, allows comparison across problems with differing objective counts (raw HV values are not comparable across dimensions). We prefer HV over metrics such as Generational Distance because it is more robust to the discontinuous Pareto Fronts common in our benchmark.
For scenarios where the ideal hypervolume is not available due to computational complexity (high number of demands or objectives), we use the raw hypervolume to compare across different algorithms or architectures for each fixed task. 

While hypervolume-based metrics can evaluate the overall quality and diversity of solutions, measuring the hypervolume becomes computationally expensive for high numbers of objectives, and may not provide the full picture in evaluating multi-objective problems ~\cite{ibrahim2024novelparetooptimalrankingmethod}, especially for PCPL. In particular, an agent may predict solutions close to the ideal Pareto Front some of the time, resulting in a high hypervolume, but the hypervolume metric does not indicate the reliability of single-shot predictions.

\subsection{Proportion of Non-Dominated Solutions (PNDS)}

To address the issue of prediction reliability, we also calculate the proportion of 
the predicted solutions from Section \ref{subsubsec:hypervolume} that are non-dominated. This metric is independent of the ideal Pareto Front solution, which provides two main advantages:

\begin{enumerate}
    \item It can be computed even when the ideal Pareto Front is intractable (e.g. many production combinations), and unlike purity~\cite{custPurity}, which scores against an external reference pooled across solvers, PNDS uses only a single policy's own predictions.
    \item High PNDS with low hypervolume can indicate the agent is trapped in local optima.
\end{enumerate}

While this metric is beneficial as a supplement to the hypervolume, it still struggles when the number of objectives is high due to sampling inefficiency and requires sampling exponentially more preferences using ~\cite{DasAndDennis1998} with increasing $N$ to ensure fair comparisons.

\subsection{Ordering Score (OS)}
The ordering score measures the rank correlation between the preference weight $w_i$ on each production $i$ and the value of the objective function $\mathbf{J}_i(\mathbf{P})$. This evaluates how well the agent obeys the preference input, instead of whether the agent generates solutions close to the Pareto Front, which is measured by the hypervolume and PNDS. For traditional multi-objective optimization algorithms, we are often concerned with generating an unordered population of solutions which approximates the Pareto Front. However, for PCPL, the ordering score becomes an important way to evaluate how well an agent follows the preference it is given. 

To compute the ordering score detailed in Algorithm \ref{alg:ordering-score} in Appendix \ref{sec:ordering-score}, 
we sample preferences that sweep along each objective and evaluate the corresponding objective values. 
An agent has performed well if the reward received in a particular objective increases with the preference weight in that component. We quantify this relationship using the Spearman Rank Correlation, normalized between 0 and 1 and averaged across all objective dimensions. The final ordering score is computed as the average Spearman Rank Correlation across all samples and all objectives.

The ordering score is a low-cost, scalable, magnitude-agnostic metric: it rewards the ordering of objective values rather than their levels. Metrics such as expected utility ~\cite{Zintgraf2015QualityAO} account for preferences only indirectly, as they remain sensitive to objective magnitudes and the choice of scalarization; OS instead isolates preference alignment via a magnitude- and scalarization-free rank correlation. Being distribution-free, the Spearman correlation is robust to outliers and complements the magnitude-focused HV. Unlike the directional-angle training signal in PD-MORL, OS evaluates alignment at inference time. We thus view it as a supplementary preference-awareness diagnostic, used alongside HV and PNDS.

\section{Results and Discussion}

\begin{figure}
    \centering
    \includegraphics[width=\linewidth]{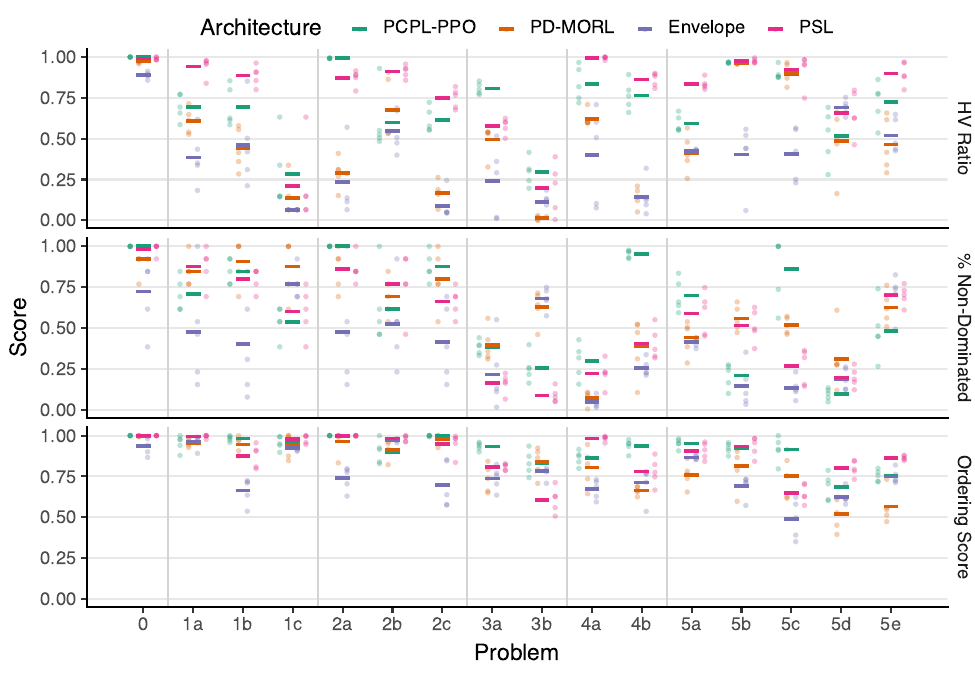}
    \caption{Performance comparison of PCPL-PPO, PD-MORL, Envelope Q-Learning and PSL policies over 5 random seeds at up to 1M steps for HV Ratio, PNDS and OS. Horizontal bars show the mean, and dots show the outcome for each seed.}
    \label{fig:problems-1-5}
\end{figure}

\begin{figure*}
    \centering
    \includegraphics[width=0.8\textwidth]{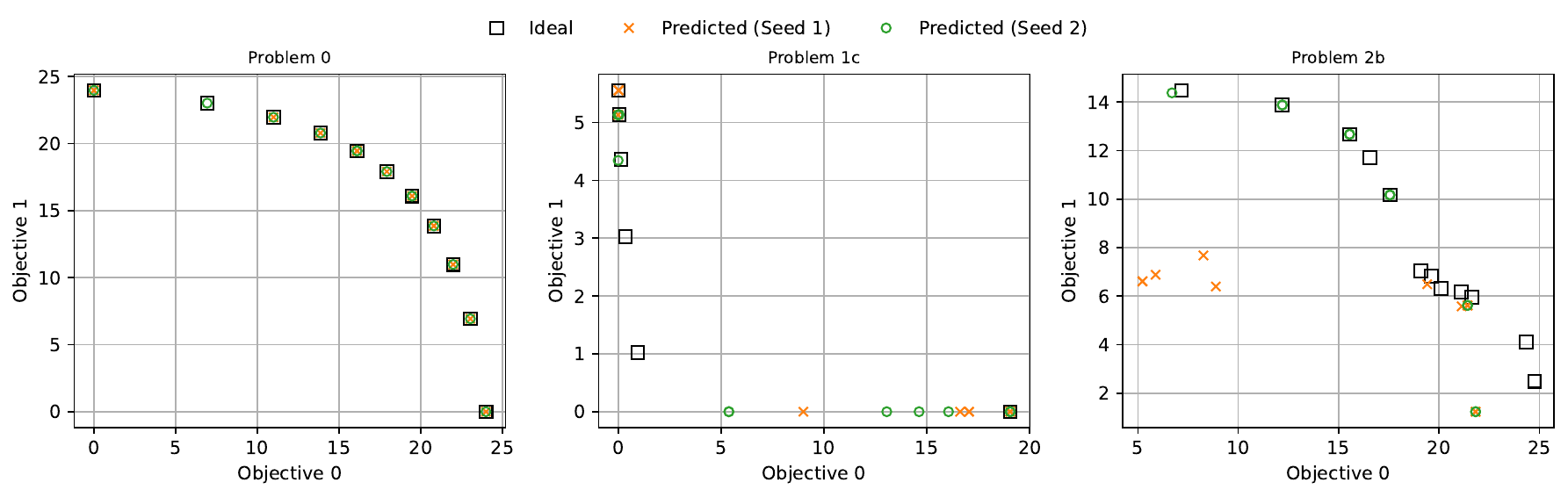}
    \caption{Selected 2-Objective Pareto Fronts for PCPL-PPO at 1M steps over 2 random seeds: Compared to the Problem 0 baseline, the PCPL-PPO agent struggles with non-convex (Problem 1c) and multi-segment (Problem 2b) Pareto Fronts.}
    \label{fig:selected-pareto-fronts}
\end{figure*}

\subsection{Main Problem Set}
Using the set of objective functions as our multi-objective reward function $\mathbf{J}$, we trained a preference-conditioned policy using Proximal Policy Optimization (PPO) on GraphAllocBench to convergence (See Figure \ref{fig:summary_figure}). We will refer to our implementation as PCPL-PPO. We then employ the methods in Section \ref{subsec:graphallocbench} to evaluate the trained model.
During training, preferences are sampled from a flat Dirichlet distribution for each rollout, and the PPO agent is trained to maximize the scalarized reward (using Smooth Tchebycheff Scalarization ~\cite{lin2024smoothtchebycheffscalarizationmultiobjective}), given the sampled preference vector and current normalized allocation matrix as input. Each rollout lasts for a total of $T$ steps. See Appendix \ref{sec:training} for more information on our PCPL-PPO implementation. Figure \ref{fig:problems-1-5} compares PCPL-PPO against three other PCPL algorithms on problem categories 0-5: PD-MORL ~\cite{basaklar2023pdmorl}, Envelope Q-Learning ~\cite{envelope} and Pareto Set Learning (PSL) ~\cite{liu2025pareto}. The benchmark's diverse problem set elicits distinct failure modes across methods, and the supplementary PNDS and OS metrics reveal preference-alignment gaps that hypervolume alone does not capture. PSL attains the strongest hypervolume across most problems, while PCPL-PPO leads on PNDS and is most robust on the hardest sparse and non-convex problems; the two are comparable on OS. Full results are in Appendix \ref{sec:results}.

These metrics are most informative where they disagree with hypervolume. On Problem 5c, PSL attains the highest hypervolume yet PCPL-PPO leads on both PNDS and ordering score; and on the five-objective problems 3a, 3b, and 4a, PSL's high-coverage sets yield few non-dominated solutions (PNDS 0.16, 0.09, 0.22). Strong hypervolume thus does not guarantee reliable, preference-aligned predictions — directly motivating PNDS and OS alongside HV.

\subsubsection{Sharp Changes in Rewards (e.g. Problems 1a-c)}
These sudden changes in reward signals make it difficult for an agent to learn a smooth approximation of the Pareto Front, and also lead to greater variance in predictions. From Figure \ref{fig:problems-1-5}, compared to the baseline (Problem 0) with a smooth objective function, Problems 1a-c report significantly higher variance and worse approximations of the Pareto Front despite sharing a dependency structure with the baseline, since sharp objective functions can result in large jumps between adjacent discrete points on the Pareto Front, which can be difficult to predict based on a smooth preference vector input.

\subsubsection{Sparse Rewards (e.g. Problem 1c)}

A PCPL agent may find it difficult to deal with sparse rewards, especially when optimizing a continuous preference vector. As example of this would be the $\mathbf{J}_i(\mathbf{P}) = \text{max}(0, \mathbf{P}_j - 5)$ floor function, where we receive no reward for Objective $i$ until we produce more than $5$ units for Demand $j$. Instead of deciding whether to a) optimize a particular objective by creating productions until the reward spikes or to b) allocate no resources to the given production, the agent might create insufficient production, such that resources are wasted but no additional reward is received. This 
results in a lower proportion of non-dominated solutions, as observed for Problem 1c in Figure \ref{fig:problems-1-5}. In Figure \ref{fig:selected-pareto-fronts}, we also observe many predicted solutions along the horizontal axis where the reward component for Objective 1 is 0. However, there should only be one Pareto-optimal solution along the horizontal axis that maximizes Objective 0.

\subsubsection{Non-convex Pareto Front (e.g. Problems 1c, 3b)}
Non-convex Pareto Fronts are difficult to represent in general for multi-objective optimization or RL problems, and are highly dependent on the scalarization function used. Smooth Tchebycheff Scalarization is more robust to non-convexity than weighted sums, but still has limitations when the Pareto front is non-convex and discontinuous, as shown in Problem 1c and 3b (See Figures \ref{fig:problems-1-5} and \ref{fig:selected-pareto-fronts}).

\subsubsection{Unbalanced Objectives (e.g. Problem 2c)}
Objectives with sparse reward signal can be difficult to optimize jointly with other concurrent objectives, which can result in failure to reach the sparser objective's extreme values and achieving a lower-quality approximation of intermediate values on the Pareto Front (See Problem 2c in Figure \ref{fig:problems-1-5}).

\subsubsection{Local Optima Traps (e.g. Problem 2b)}
When approximating a continuous Pareto front from smooth preference inputs, the PCPL agent can fail to capture certain regions of the front in the presence of local optima (Problem 2b). As shown in Figure \ref{fig:selected-pareto-fronts}, the agent collapses to local solutions and cannot reliably generalize to the global Pareto front, highlighting a key failure mode of existing approaches.

\section{Heterogeneous Graph Neural Network} \label{sec:gnn}
For problems 6a-c with a large number of resources and productions, we used a Heterogeneous Graph Neural Network (HGNN) ~\cite{DBLP:journals/corr/abs-2003-01332} to represent the complex bipartite structure of the demand-resource dependencies. Extracting global information from Graph Neural Networks to use as feature input to RL models is challenging, because global pooling methods can potentially dilute the quantity and quality of information available in the original environment. To tackle this issue, past works have used specific node embeddings ~\cite{wang2025rlhgnnreinforcementlearningdrivenheterogeneous} and node embedding concatenation ~\cite{sharma2025grapheonrlgraphneuralnetwork}; thus, we hope to show that flexible pooling methods like Mean/Max Pooling and Attention Pooling can still capture global information more effectively than MLPs. We replace the common MLP feature extractor with a HGNN-based feature extractor by using multiple Heterogeneous Graph Attention layers and performing global pooling on the node embeddings for different types of nodes separately. Each HGNN layer is comprised of several Graph Attention Networks ~\cite{veličković2018graphattentionnetworks}, one for each type of node (Demand / Resource / Unallocated). Two HGNN layers are stacked 
with residual connections in between. 

\subsection{HGNN Preference Conditioning}

To increase the preference-awareness of the policy, we incorporate preference conditioning at multiple stages within the HGNN feature extractor and training pipeline, enhancing the model's robustness to varying preference inputs.

\subsubsection{Concatenating Preferences to Features}
We concatenate the preference vector to both the initial node embeddings and the final pooled node embeddings. Additionally, residual connections between HGNN layers are employed to preserve preference information across layers throughout the forward pass.

\subsubsection{Preference-conditioned Attention Pooling}
We use preference-conditioned multi-head attention pooling, where each head computes attention scores over the nodes by jointly considering the node features and the preference vector $\mathbf{w}$. For each head, the node embeddings are transformed together with $\mathbf{w}$, weighted by the attention scores, and aggregated into a pooled representation. The outputs of all heads are then concatenated to form the final preference-aware global graph representation.

\begin{table}
    \caption{Metrics for Problem 6 (100x100 graph) using PCPL-PPO with MLP (1.58M params), HGNN+MeanMaxPool (635K params), and HGNN+AttentionPool (1.15M params) at 3M steps across 5 random seeds. The hypervolume (HV) measures overall policy performance (mean $\pm$ std). The mean scores for PNDS and Ordering Score  measure consistency and preference alignment.}
    \label{tab:ablation}
    \centering
    \begin{tabular}{lccc}
        \toprule
        \textbf{Problem 6a} & HV $\times 10^6$ & PNDS & Order\\
        \midrule
        MLP & $3.6 \pm 0.6$ & $\pmb{0.20}$ & $\pmb{0.86}$\\
        HGNN+MeanMaxPool & $9.3 \pm 2.1$ & $0.13$ & $0.72$\\
        HGNN+AttentionPool & $\pmb{14.2 \pm 2.8}$ & $0.16$ & $0.67$\\

        \midrule
        \textbf{Problem 6b} & HV $\times 10^6$ & PNDS & Order \\
        \midrule
        MLP & $4.2 \pm 0.6$ & $0.16$ & $\pmb{0.87}$\\
        HGNN+MeanMaxPool & $7.9 \pm 2.4$ & $0.11$ & $0.74$\\
        HGNN+AttentionPool & $\pmb{14.6 \pm 2.4}$ & $\pmb{0.17}$ & $0.70$\\
        \midrule
        \textbf{Problem 6c} & HV $\times 10^6$ & PNDS & Order \\
        \midrule
        MLP & $7.9 \pm 1.1$ & $0.13$ & $\pmb{0.82}$\\
        HGNN+MeanMaxPool & $25.2 \pm 3.1$ & $0.14$ & $0.63$\\
        HGNN+AttentionPool & $\pmb{33.7 \pm 4.1}$ & $\pmb{0.15}$ & $0.54$\\
        \bottomrule
    \end{tabular}
\end{table}

\subsection{MLP vs HGNN Model Comparison} \label{subsec:hgnn-comparison}
For small graphs, MLP-based models are preferred for their stability and rate of convergence. However, 
Graph Neural Networks have the potential to better generalize to large graphs with large dependency structures and eventually outperform MLPs. We investigated this behavior for Problems 6a-c 
by increasing the number of demands and resources to $100$ each, with a total of $201$ nodes (including ``Unallocated'' resource node). From Table \ref{tab:ablation}, the HGNN-based approaches achieve significantly and consistently better hypervolume scores than MLPs. The two pooling variants (Mean+Max vs. preference-conditioned Attention Pooling) span fixed versus learned aggregation. Attention Pooling attains the highest HV but the lowest OS, indicating that learned aggregation maximizes global coverage at the expense of per-objective preference alignment. Both HGNN methods have fewer parameters than the MLP-based PPO model, as the number of MLP parameters scales quadratically with the size of the allocation matrix, due to the flattening of the allocation matrix and the initial fully connected feedforward neural network. The MLP outperforms the HGNN versions on 
ordering score, which may indicate that the MLP finds more preference-aligned local solutions than the HGNN-based models, but struggles to find solutions closer to the global solution (See Section \ref{subsec:graphallocbench}).


\section{Conclusion}


We introduced \textbf{GraphAllocBench}, a flexible benchmark for preference-aware RL and combinatorial resource allocation powered by our custom \textbf{CityPlannerEnv} sandbox. GraphAllocBench's diverse problem set exposes distinct failure modes across both SOTA methods and our proposed PCPL-PPO approach, demonstrating that no single method dominates across all settings. Crucially, our supplementary metrics PNDS and Ordering Score reveal preference-alignment gaps that hypervolume alone cannot capture: a solution with lower HV may better reflect user preferences as measured by OS, underscoring the importance of multi-dimensional evaluation for PCPL. Our experiments further show that equipping PPO with an HGNN feature extractor outperforms standard MLP-based baselines on large graphs, though trade-offs in preference consistency remain an open challenge. Overall, GraphAllocBench serves as a versatile and challenging testbed for advancing research in preference-aware many-objective policy learning, and future work will extend the benchmark with richer dependency structures and environmental uncertainty to evaluate risk-aware decision-making.

\section*{Acknowledgments}

The project or effort depicted was or is sponsored by the U.S. Army Combat Capabilities Development Command -- Soldier Center under contract number W912CG-24-D-0001. The content of the information does not necessarily reflect the position or the policy of the Government, and no official endorsement should be inferred. 
Distribution Statement A. Approved for public release: distribution is unlimited.

\bibliographystyle{named}
\bibliography{references}

\appendix
\onecolumn
\appendixpage
\section{Pareto Front} \label{sec:pareto-front}

In multi-objective optimization, a solution is Pareto optimal if it cannot be improved in any objective without degrading at least one other objective. Let's consider a multi-objective maximization problem with $N$ objectives, represented by a vector function:
\begin{align*}
\mathbf{F}(x) = (f_1(x), f_2(x), \dots, f_N(x)),
\end{align*}
where $x$ is a solution in the decision space $X$.

A solution $x_1 \in X$ is said to \textbf{dominate} another solution $x_2 \in X$, denoted as $x_1 \succ x_2$, if and only if:
\begin{enumerate}
    \item $f_i(x_1) \ge f_i(x_2)$ for all objectives $i = 1, \dots, N$.
    \item $f_j(x_1) > f_j(x_2)$ for at least one objective $j \in \{1, \dots, N\}$.
\end{enumerate}

A solution $x^* \in X$ is called \textbf{Pareto optimal} if there is no other solution $x \in X$ that dominates it. The set of the objective values of all Pareto-optimal solutions form the \textbf{Pareto Front}.

\section{Hypervolume Definition} \label{sec:hv-appendix}
The hypervolume of a set of solutions generated from a particular policy is defined as follows:

\begin{align}
\text{HV}(X, \mathbf{r}) = 
\lambda \left( 
\bigcup_{\mathbf{x} \in X} 
\big( [r_0, x_0] \times \cdots \times [r_{N-1}, x_{N-1}] \big)
\right),
\end{align}
where:
\begin{itemize}
  \item $X \subset \mathbb{R}^N$ is a finite set of objective vectors (solutions in the objective space),
  \item $\mathbf{x} = (x_0, x_1, \ldots, x_{N-1}) \in X$ is an individual objective vector,
  \item $\mathbf{r} = (r_0, r_1, \ldots, r_{N-1}) \in \mathbb{R}^N$ is the reference point, chosen such that $x_i \ge r_i$ for all $i$ (maximization case),
  \item $N$ is the number of objectives,
  \item $\lambda(\cdot)$ denotes the Lebesgue measure (i.e. the volume) of the dominated region.
\end{itemize}

The reference point is standardized to be at the origin for all problems, since all objective functions are designed to be non-negative.

\section{Ordering Score Computation} \label{sec:ordering-score}
Refer to Algorithm \ref{alg:ordering-score} for Ordering Score computation.

\begin{algorithm}[h!]
\caption{Ordering Score Computation}
\label{alg:ordering-score}
\begin{algorithmic}[1]
\Require Environment $env$ with $N$ objectives, policy $\pi_\theta$, number of samples $n_{\text{samp}}$, steps $n_{\text{step}}$ in each parameter sweep
\Ensure Ordering score $\mathcal{O} \in [0,1]$

\State Generate set of preferences $\mathbf{W}$ by sweeping each objective $i$ from $0$ to $1$ in $n_{\text{step}}$ steps, 
      sampling other components uniformly via random Dirichlet samples (repeat $n_{\text{samp}}$ times).

\State Evaluate $\pi_\theta$ $\forall \mathbf{w} \in \mathbf{W}$ to obtain a list of objective vectors.

\For{each objective $i$ and repetition $j$}
    \State Extract reward sequence $\mathbf{J}_i^{(j)}$ across $n_{\text{step}}$ steps.
    \If{all values in $\mathbf{J}_i^{(j)}$ are equal} 
        \State $s_{i,j} \gets 1$
    \Else 
        \State $s_{i,j} \gets \tfrac{1}{2}\big( \text{Spearman}(\mathbf{J}_i^{(j)}, \text{sorted}(\mathbf{J}_i^{(j)})) + 1 \big)$
    \EndIf
\EndFor

\State \Return $\mathcal{O} = \tfrac{1}{N \cdot n_{\text{samp}}} \sum_{i=1}^{N} \sum_{j=1}^{n_{\text{samp}}} s_{i,j}$
\end{algorithmic}
\end{algorithm}

\section{Training Methodology} \label{sec:training}

\begin{figure*}
    \centering
    \includegraphics[width=\textwidth]{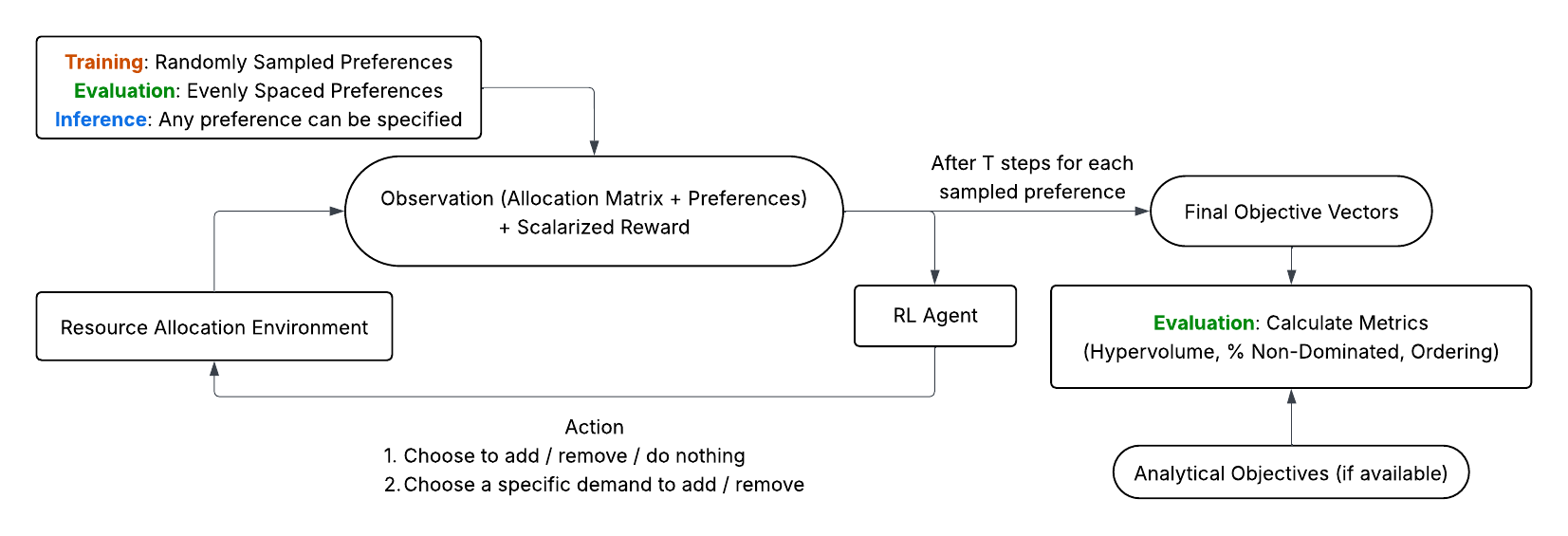}
    \caption{Training and evaluation pipeline using the CityPlannerEnv Gymnasium environment and Proximal Policy Optimization (PPO) agent from Stable Baselines3 \protect\cite{stable-baselines3}.}
    \label{fig:flowchart}
\end{figure*}

\subsection{PCPL-PPO Solution}

Using the set of objective functions as our multi-objective reward function $\mathbf{J}$, we train a preference-conditioned policy using Proximal Policy Optimization (PPO) to convergence. We then employ the methods in Section \ref{subsec:graphallocbench} to evaluate the trained model.
Figure \ref{fig:flowchart} illustrates our training and evaluation pipeline. During training, preferences are sampled from a flat Dirichlet distribution for each rollout, and the PPO agent is trained to maximize the scalarized reward based on different preferences for each rollout, given the preference vector as input. Each rollout lasts for a total of $T$ steps, which is part of the environment configuration set prior to training. 


\subsection{Normalization} \label{subsec:normalization}
Firstly, we normalized each component of the objective function to between 0 and 1, by taking the worst values of each component (i.e. the nadir point) to be at the origin (since objectives are defined to be always non-negative), and setting the best values (i.e. ideal point) to the current maximum reward obtained based on all prior rollouts. Even though using a moving ideal point may lead to some instability in training when discovering new ideal points, the agent eventually learns the optimal behavior over a large number of training steps.

\subsection{Scalarization} \label{subsec:scalarization}
Most existing RL-based scalarization methods use weighted sum scalarizations (i.e. $\text{Reward} = \mathbf{J} \cdot \mathbf{w}$) because of their simplicity. 
However, weighted-sum methods are ineffective in approximating non-convex Pareto Fronts. Therefore, we experimented with other scalarization methods, such as Tchebycheff scalarization ~\cite{tchebycheff1983} and Penalty Boundary Intersection (PBI) ~\cite{PBIDebAndJain}. We found that Smooth Tchebycheff Scalarization ~\cite{lin2024smoothtchebycheffscalarizationmultiobjective} performed the most robustly in terms of hypervolume achieved when the agent encountered highly non-convex Pareto fronts. Additionally, the Smooth Tchebycheff Scalarization includes a smoothness hyperparameter that can be tuned for different problems with varying Pareto Front landscapes.

Unlike weighted sums, methods such as the Smooth Tchebycheff Scalarization require an ideal point to compute the scalarized reward, and this information is not readily available in an RL environment. Hence we use the moving ideal point ~\cite{6615007} introduced in Section \ref{subsec:normalization} as an estimate of the true ideal point.


\subsection{Model Architecture}
\begin{figure}[h!]
    \centering
    \includegraphics[width=0.5\linewidth]{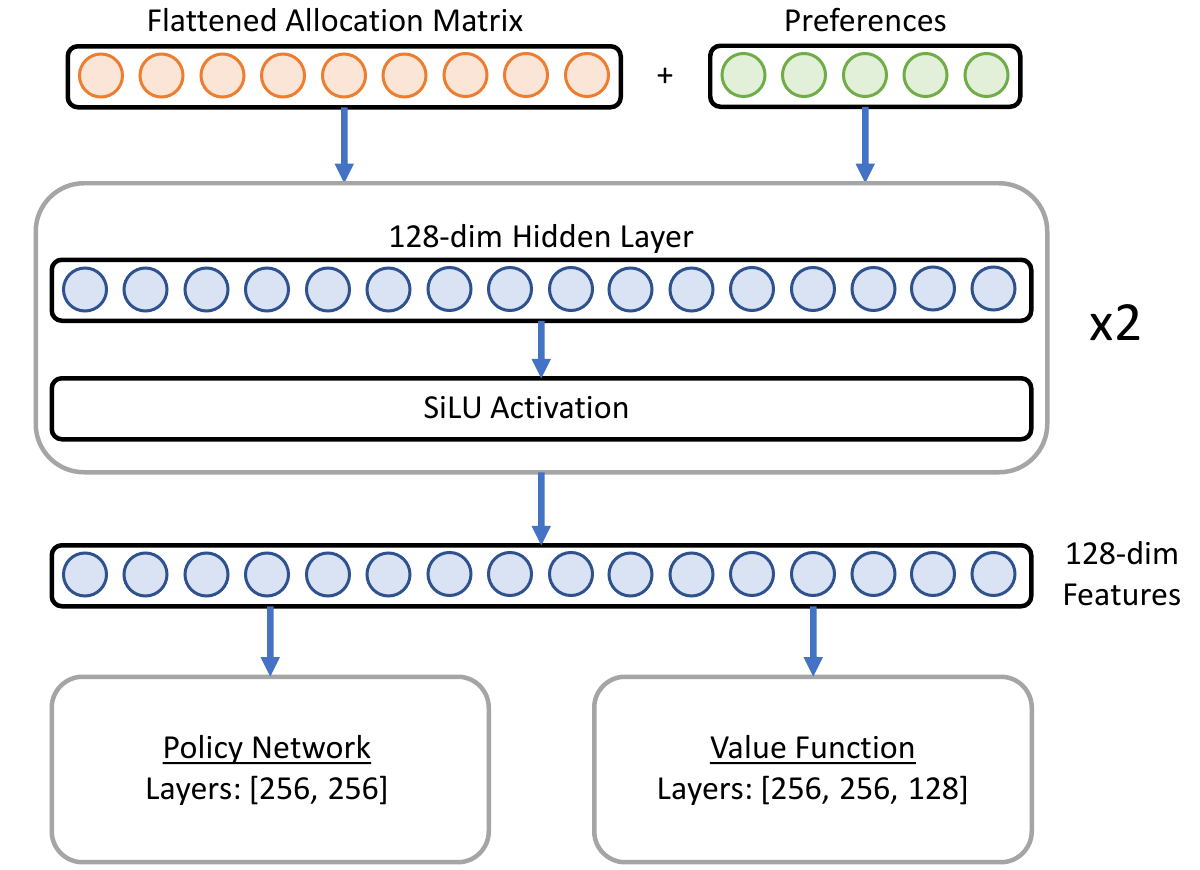}
    \caption{PPO architecture with a shared 2-layer MLP feature extractor (128 units, SiLU) for policy and value networks, implemented with Stable Baselines3.}
    \label{fig:mlp-arch}
\end{figure}

We used a Stable Baselines3 Proximal Policy Optimization (PPO) Agent with a Multi-Layer Perceptron (MLP) feature extractor (See Figure \ref{fig:mlp-arch}). This feature extractor can be easily replaced with alternate architectures such as Graph Neural Networks (Section \ref{sec:gnn}), while maintaining the same feature vector dimension. 


\section{Objective Function} \label{sec:obj-fun}
See Table \ref{tab:obj-fun} for examples of objective functions defined in GraphAllocBench using CityPlannerEnv. Each component of the objective function $\mathbf{J}_i$ can depend on any production $\mathbf{P}_j$, and comprises elementary mathematical functions such as linear, quadratic, logarithmic, logistic, Gaussian, ceiling and floor functions. We can obtain complex objective functions based on the sum of elementary functions to design particular challenges (e.g. Gaussian with ceiling, sinusoidal increasing). All objectives are non-negative (i.e. $\mathbf{J}_i = \max(f(\mathbf{P}),0)$) so that reward is always non-negative. For more detailed problem definitions on the full problem set, refer to https://github.com/jzh001/GraphAllocBench.

\begin{table*}
    \caption{Evaluation Problem Set Description}
    \label{tab:problem-descriptions}
    \begin{tabular*}{\textwidth}{@{\extracolsep{\fill}} r c c c p{0.75\textwidth}}
        \toprule
        \textbf{Problems} & $|P|$ & $N$ & \textbf{FC} & \textbf{Description} \\ 
        \midrule
        0 & 2 & 2 & Yes & Baseline simple logarithmic objective functions with smooth convex Pareto Fronts. The baseline problem uses a Fully-Connected (FC) dependency graph, where every demand depends on every resource.\\
        1a-c & 2 & 2 & Yes & Difficult objective functions, including oscillatory behavior, stationary rewards, and spikes. \\
        2a-c & 5 & 2 & Yes & More demands, which increases the action and observation spaces. \\
        3a-b & 5 & 5 & Yes & 5 objectives with sparse rewards, building on difficult objective functions from previous testcases. \\
        4a-b & 5 & 5 & No & Varied dependencies and resources, instead of Fully-Connected (FC) dependency graphs. \\
        5a-e & 5 & 3-5 & No & Testcases with dependencies, resources, and objectives sampled randomly. Extended horizon for allocating more available resources.\\
        6a-c & 100 & 5 & No & Random testcases with simple convex functions similar to baseline, but with more complex graph structures with 100 demands and 100 resources. The ideal Pareto Front is not computed due to computational complexity.\\
        \bottomrule
    \end{tabular*}
    *For more detailed problem definitions, refer to Appendix \ref{sec:obj-fun}.
\end{table*}

\begin{table*}[htbp]
\centering
\caption{Objective functions for selected problems (zero lower bound assumed for all functions).}
\label{tab:obj-fun}
\begin{tabular}{ll}
\toprule
Problem & Objective Function $\mathbf{J}$ \\
\midrule
\multirow{2}{*}{0} & $\mathbf{J}_0 = 10 \log (\mathbf{P}_0 + \epsilon + 1)$ \\
                           & $\mathbf{J}_1 = 10 \log (\mathbf{P}_1 + \epsilon + 1)$\\
\midrule
\multirow{2}{*}{1a} & $\mathbf{J}_0 = \min(0.6 \mathbf{P}_0^2, 18) + \max (5 \exp(-0.9(\mathbf{P}_1-7)^2), 0.1)$ \\
                           & $\mathbf{J}_1 = -1.5 \mathbf{P}_0^2 + 0.8\mathbf{P}_0 + 12 + 5\exp(-0.1(\mathbf{P}_1-3)^2) + 0.2\mathbf{P}_1\log(2\mathbf{P}_1 + \epsilon + 1)$ \\
\midrule
\multirow{2}{*}{1b} & $\mathbf{J}_0 = 0.5\mathbf{P}_0 + 3 + 3 \sin (3\mathbf{P}_0) - \log (1.5\mathbf{P}_1 + \epsilon + 2) + 2.5\exp(-1.2(\mathbf{P}_1-7)^2)$\\
                           & $\mathbf{J}_1 = -1.5\log(5.5\mathbf{P}_0 + \epsilon + 1) + 9\exp(-0.5(\mathbf{P}_0 - 2)^2) + 5 + (0.5\mathbf{P}_1 + 1.2)\sin(0.9\mathbf{P}_1 + 0.6)$\\
\midrule
\multirow{2}{*}{1c} & $\mathbf{J}_0 = -\mathbf{P}_0 + \frac{20}{1 + \exp(\mathbf{P}_1 - 3)}$\\
                           & $\mathbf{J}_1 = \frac{1}{1+\exp(\mathbf{P}_0 - 6)}  + 5 - \frac{15}{1+\exp(0.7(\mathbf{P}_1 - 5))}$\\
\bottomrule
\end{tabular}
\end{table*}

\section{Results and Comparison with PD-MORL, Envelope Q-Learning, and PSL}
\label{sec:results}

We ran PD-MORL~\cite{basaklar2023pdmorl} (MO-DDQN-HER), Envelope Q-Learning~\cite{envelope}, and Pareto Set Learning (PSL)~\cite{liu2025pareto} on all GraphAllocBench problems and compare against our PPO-based PCPL (PCPL-PPO) in Table~\ref{tab:pdmorl_vs_pcpl}. For each baseline we conducted a per-problem random hyperparameter search (e.g. learning rate, discount factor, batch size, soft-update rate, replay buffer size, exploration schedule, and training step budget) and used the best configuration found per problem to train 5 independent seeds at up to 1M steps.

\setlength{\tabcolsep}{4pt}
\begin{longtable}{llccc}
\caption{Comparison of PD-MORL, Envelope Q-Learning, PSL, and PCPL-PPO across all benchmark problems (mean $\pm$ std, $n{=}5$ seeds). The baselines use per-problem hyperparameters selected via random search. Bold indicates the best result per metric per problem.}
\label{tab:pdmorl_vs_pcpl}\\
\toprule
\textbf{Problem} & \textbf{Method} & \textbf{Norm.\ HV} $\uparrow$ & \textbf{PNDS} $\uparrow$ & \textbf{Ordering} $\uparrow$ \\
\midrule
\endfirsthead
\caption[]{Comparison of PD-MORL, Envelope Q-Learning, PSL, and PCPL-PPO across all benchmark problems (continued).}\\
\toprule
\textbf{Problem} & \textbf{Method} & \textbf{Norm.\ HV} $\uparrow$ & \textbf{PNDS} $\uparrow$ & \textbf{Ordering} $\uparrow$ \\
\midrule
\endhead
\midrule
\multicolumn{5}{r}{\textit{Continued on next page}} \\
\endfoot
\bottomrule
\endlastfoot
\multirow{4}{*}{0} & PD-MORL   & $0.973 \pm 0.012$ & $0.923 \pm 0.094$ & $0.997 \pm 0.003$ \\
                    & Envelope  & $0.889 \pm 0.020$ & $0.723 \pm 0.222$ & $0.937 \pm 0.054$ \\
                    & PSL       & $0.990 \pm 0.009$ & $0.985 \pm 0.034$ & $\mathbf{1.000 \pm 0.000}$ \\
                    & PCPL-PPO  & $\mathbf{1.000 \pm 0.000}$ & $\mathbf{1.000 \pm 0.000}$ & $\mathbf{1.000 \pm 0.000}$ \\
\midrule
\multirow{4}{*}{1a} & PD-MORL   & $0.608 \pm 0.078$ & $0.846 \pm 0.094$ & $0.956 \pm 0.023$ \\
                    & Envelope  & $0.385 \pm 0.159$ & $0.477 \pm 0.333$ & $0.961 \pm 0.041$ \\
                    & PSL       & $\mathbf{0.943 \pm 0.059}$ & $\mathbf{0.877 \pm 0.117}$ & $\mathbf{0.995 \pm 0.010}$ \\
                    & PCPL-PPO  & $0.696 \pm 0.079$ & $0.708 \pm 0.100$ & $0.954 \pm 0.047$ \\
\midrule
\multirow{4}{*}{1b} & PD-MORL   & $0.439 \pm 0.126$ & $\mathbf{0.908 \pm 0.138}$ & $0.945 \pm 0.052$ \\
                    & Envelope  & $0.459 \pm 0.246$ & $0.400 \pm 0.324$ & $0.665 \pm 0.080$ \\
                    & PSL       & $\mathbf{0.887 \pm 0.061}$ & $0.800 \pm 0.069$ & $0.876 \pm 0.069$ \\
                    & PCPL-PPO  & $0.696 \pm 0.122$ & $0.846 \pm 0.077$ & $\mathbf{0.982 \pm 0.017}$ \\
\midrule
\multirow{4}{*}{1c} & PD-MORL   & $0.136 \pm 0.118$ & $\mathbf{0.877 \pm 0.177}$ & $0.942 \pm 0.075$ \\
                    & Envelope  & $0.065 \pm 0.000$ & $0.769 \pm 0.094$ & $0.922 \pm 0.015$ \\
                    & PSL       & $0.211 \pm 0.239$ & $0.600 \pm 0.148$ & $\mathbf{0.980 \pm 0.025}$ \\
                    & PCPL-PPO  & $\mathbf{0.282 \pm 0.213}$ & $0.538 \pm 0.094$ & $0.955 \pm 0.040$ \\
\midrule
\multirow{4}{*}{2a} & PD-MORL   & $0.290 \pm 0.093$ & $0.862 \pm 0.126$ & $0.965 \pm 0.074$ \\
                    & Envelope  & $0.235 \pm 0.205$ & $0.477 \pm 0.333$ & $0.739 \pm 0.071$ \\
                    & PSL       & $0.873 \pm 0.047$ & $0.862 \pm 0.084$ & $0.999 \pm 0.001$ \\
                    & PCPL-PPO  & $\mathbf{0.992 \pm 0.002}$ & $\mathbf{1.000 \pm 0.000}$ & $\mathbf{1.000 \pm 0.000}$ \\
\midrule
\multirow{4}{*}{2b} & PD-MORL   & $0.676 \pm 0.191$ & $0.692 \pm 0.122$ & $0.911 \pm 0.056$ \\
                    & Envelope  & $0.548 \pm 0.126$ & $0.523 \pm 0.257$ & $0.972 \pm 0.010$ \\
                    & PSL       & $\mathbf{0.912 \pm 0.038}$ & $\mathbf{0.769 \pm 0.188}$ & $\mathbf{0.983 \pm 0.018}$ \\
                    & PCPL-PPO  & $0.599 \pm 0.187$ & $0.615 \pm 0.224$ & $0.902 \pm 0.070$ \\
\midrule
\multirow{4}{*}{2c} & PD-MORL   & $0.169 \pm 0.070$ & $0.800 \pm 0.177$ & $0.977 \pm 0.030$ \\
                    & Envelope  & $0.087 \pm 0.088$ & $0.415 \pm 0.235$ & $0.698 \pm 0.141$ \\
                    & PSL       & $\mathbf{0.748 \pm 0.061}$ & $0.662 \pm 0.088$ & $0.950 \pm 0.066$ \\
                    & PCPL-PPO  & $0.614 \pm 0.075$ & $\mathbf{0.877 \pm 0.088}$ & $\mathbf{0.999 \pm 0.003}$ \\
\midrule
\multirow{4}{*}{3a} & PD-MORL   & $0.495 \pm 0.094$ & $\mathbf{0.395 \pm 0.102}$ & $0.737 \pm 0.083$ \\
                    & Envelope  & $0.240 \pm 0.221$ & $0.216 \pm 0.209$ & $0.738 \pm 0.071$ \\
                    & PSL       & $0.578 \pm 0.048$ & $0.162 \pm 0.059$ & $0.809 \pm 0.020$ \\
                    & PCPL-PPO  & $\mathbf{0.807 \pm 0.035}$ & $0.381 \pm 0.043$ & $\mathbf{0.934 \pm 0.023}$ \\
\midrule
\multirow{4}{*}{3b} & PD-MORL   & $0.013 \pm 0.012$ & $0.627 \pm 0.101$ & $\mathbf{0.838 \pm 0.088}$ \\
                    & Envelope  & $0.112 \pm 0.060$ & $\mathbf{0.680 \pm 0.067}$ & $0.782 \pm 0.047$ \\
                    & PSL       & $0.197 \pm 0.156$ & $0.087 \pm 0.044$ & $0.607 \pm 0.078$ \\
                    & PCPL-PPO  & $\mathbf{0.295 \pm 0.083}$ & $0.257 \pm 0.087$ & $0.833 \pm 0.075$ \\
\midrule
\multirow{4}{*}{4a} & PD-MORL   & $0.621 \pm 0.049$ & $0.069 \pm 0.043$ & $0.804 \pm 0.117$ \\
                    & Envelope  & $0.399 \pm 0.292$ & $0.048 \pm 0.035$ & $0.672 \pm 0.059$ \\
                    & PSL       & $\mathbf{0.994 \pm 0.009}$ & $0.221 \pm 0.080$ & $\mathbf{0.982 \pm 0.016}$ \\
                    & PCPL-PPO  & $0.835 \pm 0.107$ & $\mathbf{0.297 \pm 0.107}$ & $0.862 \pm 0.045$ \\
\midrule
\multirow{4}{*}{4b} & PD-MORL   & $0.141 \pm 0.061$ & $0.386 \pm 0.179$ & $0.663 \pm 0.026$ \\
                    & Envelope  & $0.143 \pm 0.105$ & $0.255 \pm 0.052$ & $0.713 \pm 0.102$ \\
                    & PSL       & $\mathbf{0.863 \pm 0.042}$ & $0.404 \pm 0.098$ & $0.780 \pm 0.082$ \\
                    & PCPL-PPO  & $0.764 \pm 0.088$ & $\mathbf{0.953 \pm 0.023}$ & $\mathbf{0.938 \pm 0.037}$ \\
\midrule
\multirow{4}{*}{5a} & PD-MORL   & $0.414 \pm 0.117$ & $0.442 \pm 0.105$ & $0.760 \pm 0.075$ \\
                    & Envelope  & $0.425 \pm 0.009$ & $0.415 \pm 0.029$ & $0.866 \pm 0.012$ \\
                    & PSL       & $\mathbf{0.834 \pm 0.033}$ & $0.589 \pm 0.129$ & $0.907 \pm 0.050$ \\
                    & PCPL-PPO  & $0.593 \pm 0.052$ & $\mathbf{0.699 \pm 0.100}$ & $\mathbf{0.952 \pm 0.028}$ \\
\midrule
\multirow{4}{*}{5b} & PD-MORL   & $0.962 \pm 0.005$ & $\mathbf{0.558 \pm 0.077}$ & $0.815 \pm 0.132$ \\
                    & Envelope  & $0.404 \pm 0.199$ & $0.145 \pm 0.130$ & $0.692 \pm 0.068$ \\
                    & PSL       & $\mathbf{0.976 \pm 0.013}$ & $0.514 \pm 0.100$ & $\mathbf{0.934 \pm 0.060}$ \\
                    & PCPL-PPO  & $0.966 \pm 0.004$ & $0.209 \pm 0.075$ & $0.922 \pm 0.032$ \\
\midrule
\multirow{4}{*}{5c} & PD-MORL   & $0.903 \pm 0.062$ & $0.516 \pm 0.050$ & $0.751 \pm 0.132$ \\
                    & Envelope  & $0.405 \pm 0.161$ & $0.132 \pm 0.064$ & $0.488 \pm 0.117$ \\
                    & PSL       & $\mathbf{0.924 \pm 0.099}$ & $0.266 \pm 0.109$ & $0.647 \pm 0.056$ \\
                    & PCPL-PPO  & $0.898 \pm 0.039$ & $\mathbf{0.859 \pm 0.202}$ & $\mathbf{0.916 \pm 0.076}$ \\
\midrule
\multirow{4}{*}{5d} & PD-MORL   & $0.484 \pm 0.196$ & $\mathbf{0.312 \pm 0.180}$ & $0.521 \pm 0.098$ \\
                    & Envelope  & $\mathbf{0.692 \pm 0.047}$ & $0.188 \pm 0.066$ & $0.625 \pm 0.047$ \\
                    & PSL       & $0.658 \pm 0.135$ & $0.197 \pm 0.060$ & $\mathbf{0.800 \pm 0.047}$ \\
                    & PCPL-PPO  & $0.516 \pm 0.166$ & $0.094 \pm 0.034$ & $0.686 \pm 0.080$ \\
\midrule
\multirow{4}{*}{5e} & PD-MORL   & $0.464 \pm 0.164$ & $0.625 \pm 0.098$ & $0.565 \pm 0.099$ \\
                    & Envelope  & $0.520 \pm 0.105$ & $\mathbf{0.703 \pm 0.125}$ & $0.749 \pm 0.044$ \\
                    & PSL       & $\mathbf{0.900 \pm 0.070}$ & $0.699 \pm 0.060$ & $\mathbf{0.864 \pm 0.014}$ \\
                    & PCPL-PPO  & $0.723 \pm 0.126$ & $0.483 \pm 0.171$ & $0.754 \pm 0.036$ \\
\end{longtable}

The results illustrate how GraphAllocBench differentiates algorithm behavior across diverse problem settings. PSL achieves the highest normalized hypervolume on 10 of 16 problems and beats PCPL-PPO head-to-head on hypervolume on 11 of 16, making it the strongest method for overall Pareto-front coverage. PCPL-PPO instead leads on PNDS (7 of 16 problems), and the two methods are comparable on ordering score, each leading on 8 problems (tying on Problem 0). Notably, PCPL-PPO remains the most robust method on the hardest sparse and non-convex problems (1c, 3a, 3b), where it still attains the best hypervolume. PD-MORL does not lead hypervolume on any problem but stays competitive on PNDS (5 wins), particularly on the sharp and oscillatory objectives of Problems 1b and 1c. Envelope Q-Learning leads hypervolume only on Problem 5d, and generally underperforms the other methods across the broader problem set, particularly on problems with sparse rewards (Problems 1c, 3a) and high-dimensional observations (5b–5c).

As discussed in Section 5.1, these supplementary metrics expose preference-alignment weaknesses on the five-objective problems (3a, 3b, 4a) and Problem 5c that hypervolume alone would miss; full per-problem values appear in Table 4.

The benchmark's diversity also reveals where all methods struggle most. On the hardest non-convex and sparse problems (1c, 3b), every method attains low hypervolume (best HV $0.282$ and $0.295$, both from PCPL-PPO), reflecting the difficulty of recovering discontinuous Pareto fronts. The value-based baselines PD-MORL and Envelope Q-Learning see substantially degraded performance on problems with sparse rewards, complex dependency structures, or unbalanced objectives (Problems 2c, 3a, 3b, 4b), whereas PSL and PCPL-PPO remain comparatively strong. These results demonstrate that GraphAllocBench's varied problem set and complementary metrics can expose distinct failure modes in preference-conditioned policy learning.

We note that these comparisons are based on five seeds with substantial variance on the hardest problems (e.g. Problem 1c, 0.282 ± 0.213 for PCPL-PPO), so per-problem head-to-head counts should be read as trends rather than statistically separated rankings.

\section{Effect of Increasing Number of Objectives}

\begin{figure}
    \centering
    \includegraphics[width=0.5\linewidth]{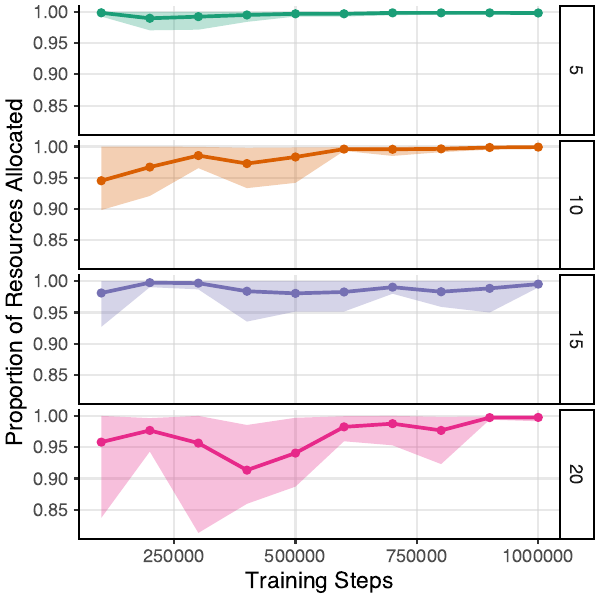}
    \caption{Performance over different number of objectives (5, 10, 15, 20) for a simple increasing objective function, across 5 random seeds.  The points show the mean performance, and the shaded regions show the range over 5 seeds.
    }
    \label{fig:logistic-more-obj}
\end{figure}


We study the effect of increasing the number of objectives (Figure \ref{fig:logistic-more-obj}) by modeling
each objective as dependent on a unique production (i.e. $N = |\mathbf{P}|$) and using objective functions that monotonically increase in each production. As the true Pareto Front is computationally intensive to calculate for high-dimensional objectives, we use the ratio of the quantity of allocated resources to total resources available to estimate the proximity of the predicted solutions to the true Pareto Front; 
intuitively, if objective functions are always monotonically increasing, it is always better to use more resources to get more reward, regardless of preference. Hence, the Pareto Front consists of outcomes that use all available resources in the environment. In Figure \ref{fig:logistic-more-obj}, results suggest that the agent maintains relatively stable performance even as the objective space dimensionality quadruples for simple objective functions, given sufficient model capacity and number of training steps. Additionally, a relatively high 20-objective ordering score ($0.89$) can still be reached at 1M steps. This restricted setup enables us to conclude that increasing the number of objectives (and productions) with a simple objective function and dependency setup may not necessarily lead to a significant reduction in overall performance, as measured by the proportion of allocated resources as an estimate of the closeness to the ideal solution. This result concurs with the findings of similar tests ~\cite{Ishibuchi2016PerformanceCO} conducted using evolutionary methods for multi-objective optimization on benchmarks such as DTLZ.

\section{Ordering Score Sensitivity}

\begin{figure}
    \centering
    \subfloat[Problem 5d]{
        \includegraphics[width=0.49\linewidth]{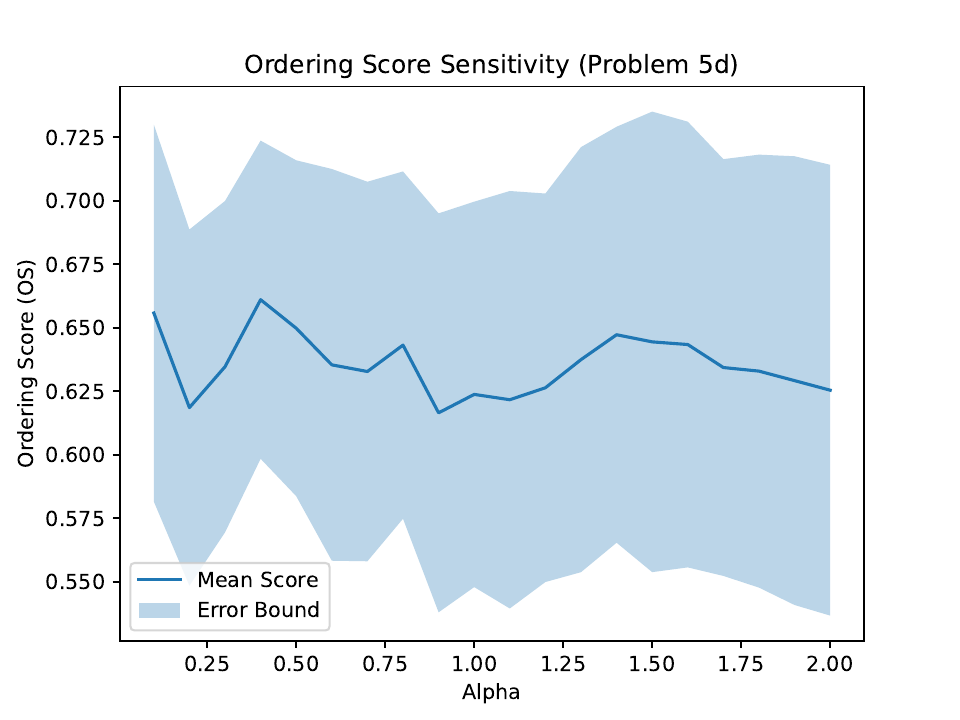}
        \label{fig:order-sensitivity-5d}
    }
    \hfill
    \subfloat[Problem 5e]{
        \includegraphics[width=0.49\linewidth]{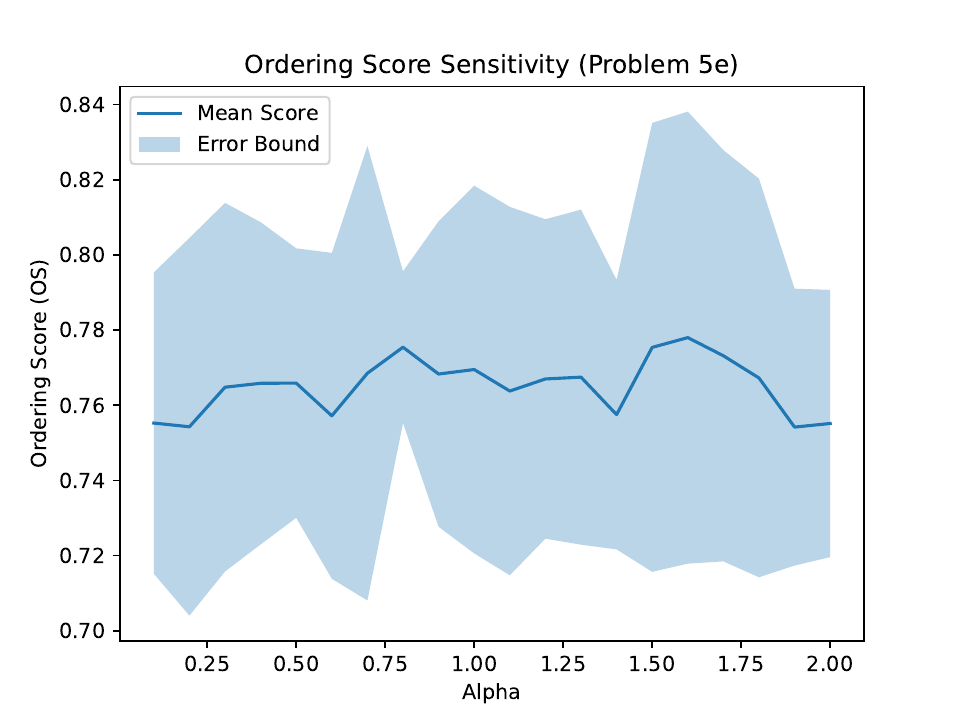}
        \label{fig:order-sensitivity-5e}
    }
    \caption{Order Sensitivity (Mean + Standard Deviation) across Dirichlet Distribution $\alpha$ parameter.}
    \label{fig:order-sensitivity}
\end{figure}

To test the sensitivity of the ordering score, we calculated the ordering score for Problems 5d and 5e across 5 random seeds with different sampling patterns by varying values of $\alpha$ for the Dirichlet Distribution sampling in Algorithm \ref{alg:ordering-score}. For simple problems (e.g. Problem 0), the ordering score remained relatively constant. For more complex problems, ordering scores may fluctuate with different sampling methods. To investigate the sensitivity of Ordering Score with respect to sampling, Problem 5d and 5e were selected for their higher number of objectives (5), varied dependencies and low ordering score. The variation in Ordering Score is shown in Figures \ref{fig:order-sensitivity-5d} and \ref{fig:order-sensitivity-5e}.

In Figures \ref{fig:order-sensitivity-5d} and \ref{fig:order-sensitivity-5e}, although there are fluctuations in ordering score due to the variations in policies obtained, the mean ordering score remains relatively stable over different alpha parameters for the Dirichlet distribution, indicating resilience over different sampling methods. 

\end{document}